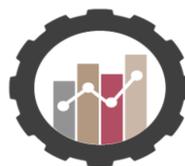

# A Combined Approach To Detect Key Variables In Thick Data Analytics

Giovanni Antonelli[1], Rosa Arboretti Giancristofaro[2], Riccardo Ceccato[3], Paolo Centomo[1], Luca Pegoraro[3], Luigi Salmaso[3*] and Marco Zecca[1]

[1)] Department of Chemical Sciences, University of Padova, Padua, Italy.
[2)] Department of Civil, Environmental and Architectural Engineering, University of Padova, Padua, Italy.
[3)] Department of Management and Engineering, University of Padova, Vicenza, Italy.

[*]Corresponding author - e-mail address: luigi.salmaso@unipd.it

**ABSTRACT** – In machine learning one of the strategic tasks is the selection of only significant variables as predictors for the response(s). In this paper an approach is proposed which consists in the application of permutation tests on the candidate predictor variables in the aim of identifying only the most informative ones. Several industrial problems may benefit from such an approach, and an application in the field of chemical analysis is presented. A comparison is carried out between the approach proposed and Lasso, that is one of the most common alternatives for feature selection available in the literature.

## 1. INTRODUCTION

Feature selection is a critical step in data preparation before machine learning modelling. Several machine learning methods have already been developed which can perform feature selection such as Lasso or tree-based methods, namely decision trees and random forests. In this paper an alternative approach is proposed which is based on NonParametric Combination (NPC) procedures [1].

The relevance of feature selection is mainly dictated by the risk of overfitting and to achieve model simplification and diminish computational time. Furthermore, in industrial environments, feature selection may lead to additional savings such as cost savings in sensors for equipment monitoring.

Several fields are affected by the problem of high dimensionality of data, therefore methods for feature selection can find application in many diverse areas. Condition monitoring of industrial equipment is one of these areas: at the present, it is not unusual to measure tens or hundreds of parameters in a device or equipment, and the more complex is the system considered the higher is the number of sensors installed on it. The recent topics of Industry 4.0 and predictive maintenance are an additional boost for the increasing number of features measured [2]. Prediction of air pollution is another topic in which feature selection plays a relevant role. Datasets for air pollution forecasting tend to have a significant number of variables, not only carrying information on the atmospheric conditions but also including covariates such as seasonality, weekday and location [3]. The techniques employed in analytical chemistry for qualitative and quantitative assessments in several sectors including chemical, petrochemical, pharmaceutical, environmental, and agricultural readily produce datasets containing hundreds of data points, hence can be used for the prediction of a dependent response based on hundreds of variables [4]. In this context, an appropriate selection of the significant range for the collection of experimental data points leads to better accuracy and performance of the analytical method together with a simpler model to use and interpret.

The approach proposed of adopting NPC procedures for selection of informative variables is here validated by its application in a problem of analyte detection in a mixture.

## 2. METHODOLOGY

The problem on hand deals with the recognition of analyte concentration in a mixture. A Design of Experiments (DOE) study is planned to obtain 250 samples that are then split in 75% for training and 25% for model validation. Each of the samples is obtained as a mixture of 6 components that are dissolved in a solvent and each of the components is split into 3 levels of concentration. For each of the 250 combinations 5 repetitions of the analytical measure are carried out in such a way that 130 data points are collected per single repetition. This results in a multivariate regression problem in which 130 variables are available for the prediction of 6 response variables. The goal is to develop a model which can reliably detect the composition of the mixture. To this end, the range of the datapoint collection is selected by means of an NPC approach.

Variable selection is carried out on the training data. The average value for the 5 repeated measures is calculated for each one of the 130 variables of each sample, and a pairwise comparison is performed (using the difference in means as test statistic) for all the samples. A permutation approach is adopted in order to perform the testing procedure and p-values are calculated accordingly [1]. This results in 17391 pairwise comparisons, and the familywise error rate is controlled by means of the false discovery rate (FDR) correction [5]. The corrected p-values are evaluated at a significance level α=0.05, and as such a certain number of significances are found for each data point. At this level, cutoffs must be found for the number of significances to determine how many variables to select. Some relevant cutoffs have been selected to the scope of this paper.

Lasso is used as a baseline for assessment of the NPC procedure proposed. Because of the multivariate nature of the problem on hand, a variable is included if it is selected by Lasso for at least one of the responses. In this setting, the multivariate Lasso selects 25 variables. After variable selection is executed, machine learning models are fit to the remaining variables of the training set and the predictive ability of the algorithms is assessed by application on the vaildation set. The validation set error is used as a means to evaluate the goodness of the subset of variables included by the different approaches.

## 3. RESULTS AND DISCUSSION

A comparison of the set of variables selected by the NPC approach and Lasso shows that the first tends to select bands of data points, while the latter tends to select data points that are rather sparse (Fig. 1). This behavior is imputable to the different rationale that drives the selection in the two cases. In presence of correlated predictors Lasso tends to pick one variable and discard the others. On the other hand, the proposed NPC approach picks all the most significant variables (according to the chosen cutoff) even if they are strongly correlated. Since in the present setting adjacent variables are more correlated than distant variables, it is no surprise that the two methods behave differently.

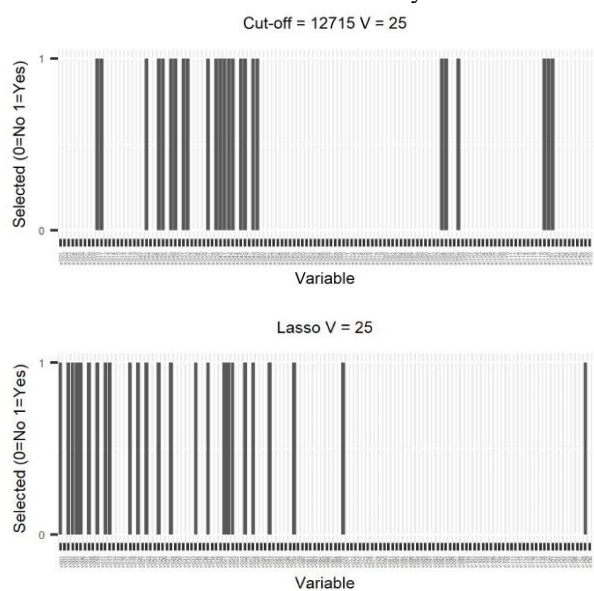

Figure 1 Variables selected by the NPC approach (top) and Lasso (bottom)

Ridge regression is selected as algorithm for machine learning modelling. The performance of the method is reported in Figure 2. The selection of only a subset of variables leads to a decrease of the predictive ability of the algorithms. The NPC approach outperforms Lasso when the number of variables selected is similar. The flexibility given by the choice of the cutoff in the NPC approach can further reduce the mean absolute error by including additional variables. Furthermore, the ability of the NPC approach to select bands of correlated variables may represent an advantage e.g. in the selection of a sensor for measuring such variables.

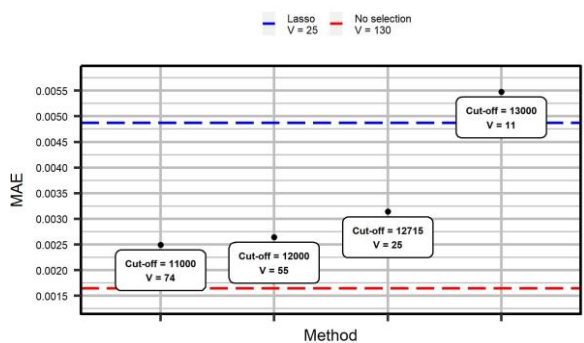

Figure 2 Validation set mean absolute error (MAE) for Ridge regression for different variable selection strategies

## 4. CONCLUSIONS

In this paper a novel strategy for feature selection based on permutation tests has been discussed and applied to a problem of chemical analysis. The performances of the approach proposed are compared to Lasso and used for model validation of a regression technique.

## 5. REFERENCES

[1] F. Pesarin, and L. Salmaso, *Permutation Tests for Complex Data: Theory, Applications and Software*, Chichester, UK: John Wiley & Sons; 2010.

[2] T. P. Carvalho, F. A.A.M.N. Soares, R. Vita, R. da P. Francisco, J. P. Basto, S. G.S. Alcalá, "A systematic literature review of machine learning methods applied to predictive maintenance," *Computers & Industrial Engineering*, vol. 137, 2019.

[3] K. Siwek, S. Osowski, "Data mining methods for prediction of air pollution," *International Journal of Applied Mathematics and Computer Science*, vol. 26, no. 2, pp. 467-478, 2016.

[4] R. M. Balabin, S. V. Smirnov, "Variable selection in near-infrared spectroscopy: Benchmarking of feature selection methods on biodiesel data," *Analytica Chimica Acta*, vol. 692, no. 1–2, pp. 63-72, 2011.

[5] Y. Benjamini, and Y. Hochberg, "Controlling the False Discovery Rate: A Practical and Powerful Approach to Multiple Testing," Journal of the Royal Statistical Society. Series B (Methodological), vol. 57, no. 1, pp. 289–300, 1995.